\documentclass{ecai} 

\usepackage{latexsym}
\usepackage{amssymb}
\usepackage{amsmath}
\usepackage{amsthm}
\usepackage{mathtools}
\usepackage{stmaryrd}
\usepackage{scalerel}
\usepackage{booktabs}
\usepackage{enumitem}
\usepackage{graphicx}
\usepackage{color}
\usepackage{placeins}

\usepackage{todonotes} 

\newcommand{\BibTeX}{B\kern-.05em{\sc i\kern-.025em b}\kern-.08em\TeX}

\def\ivBrack#1{\left\llbracket{#1}\right\rrbracket} 
\DeclareMathOperator*{\concat}{\scalerel*{\Vert}{\sum}}

\begin{document}


\begin{frontmatter}


\paperid{123} 


\title {A dual ensemble classifier used to recognise  contaminated multi-channel EMG and MMG signals in the control of upper limb bioprosthesis}


\author[A]{\fnms{Pawel}~\snm{Trajdos}\orcid{0000-0002-4337-6847}\thanks{Corresponding Author. Email: pawel.trajdos@pwr.edu.pl}}
\author[A]{\fnms{Marek}~\snm{Kurzynski}\orcid{0000-0002-0401-2725}} 

\address[A]{Wroclaw University of Science and Technology, Wybrzeze Wyspianskiego 27, 50-370,Wroclaw, Poland}


\begin{abstract}
Myopotential pattern recognition to decode the intent of the user is the most advanced approach to controlling a powered bioprosthesis.
Unfortunately, many factors make this a difficult problem and achieving acceptable recognition quality in real-word conditions is a serious challenge.
The aim of the paper is to develop a recognition system that will mitigate factors related to multimodality and multichannel recording of biosignals and their high susceptibility to contamination. The proposed method involves the use of two co-operating multiclassifier systems.
The first system is composed of one-class classifiers related to individual electromyographic (EMG) and mechanomyographic (MMG) biosignal recording channels, and its task is to recognise contaminated channels.
The role of the second system is to recognise the class of movement resulting from the patient's intention. The ensemble system consists of base classifiers using the representation (extracted features) of biosignals from different channels. The system uses a dynamic selection mechanism, eliminating those base classifiers that are associated with biosignal channels that are recognised by the one-class ensemble system as being contaminated.
Experimental studies were conducted using signals from an able-bodied person with simulation of amputation. The results obtained allow us to reject the null hypothesis that the application of the dual ensemble foes not lead to improved classification quality.

\end{abstract}


\end{frontmatter}
\flushbottom 

\section{Introduction}\label{sec:Introduction}

The loss of a hand significantly worsens the quality of a human’s life.  Although hand transplantology is making great progress, it is still not a widely used medical procedure \cite{Kay2023}. An alternative is to equip the subject with a so-called bionic upper limb prosthesis, i.e. a prosthesis controlled by biological signals.
In bioprosthetic control, the subject's intention to move the prosthesis is encoded in the contractions of the residual stump muscles that are still under the control of the motor cortex. Muscle activity generates electromiographic (EMG)  and mechanomiographic (MMG) biosignals that can be recorded from the body surface and transferred to the prosthesis.  Then, the recorded biosignals, processed in accordance with the decision-making  control  algorithm,  constitute the basis for the operation of the kinematic controler. The task of the kinematic controller is to control the DoFs of the prosthesis in such a way that the movement of the mechanical structure (trajectory of movement) ends with the posture of the prosthesis (grip  or manipulation) consistent with the intention of the subject.

During the past few decades, there has been a tremendous development in pattern recognition-based control methods for bionic upper limb prostheses.
In this approach, the patient's intention denotes the object being recognised, the recorded biosignals constitute a formal representation of the object, and the type of movement of the prosthesis is the class label.
Numerous papers covering a wide range of topics, such as the use of various classification models and feature engineering procedures (feature extraction and feature dimensionality reduction), are summarised in synthetic review publications and comprehensive comparative analyses \cite{Scheme2011,Parajuli2019,Chen2023}.
The effectiveness of complex classification methods and recognition systems with different structures and decision schemes was also investigated in this application area~\cite{Kurzynski2016,Trajdos2024,Akbulut2022}. 
Although pattern recognition-based upper limb prosthesis control is considered the most mature and advanced method, this does not translate into applications in commercial bioprostheses.
Conventional control methods still dominate here, such as a simple on-off scheme with possible manual or semi-automatic switching between prosthesis DoFs or an extended proportional control method by measuring the average amplitude of biosignals \cite{Fougner2012}.
This is because the task of pattern recognition to control a bionic prosthetic hand is a difficult problem. There are many factors that make it difficult to obtain an acceptable classification quality in real conditions, allowing the amputee to use the prosthesis in everyday activities. The most important of these factors are \cite{Hakonen2015, Kyranou2018}:
\begin{enumerate}
\item \label{itm:fac:1} The multimodal  nature of the formal representation of an object (typically: multi-channel EMG and MMG biosignals, less frequently: other biosignals (EEG),  data from extra sensors like accelerometers, magnetometers, gyroscopes, scene images from cameras),

\item \label{itm:fac:2} Difficulties in obtaining a sufficient set of training data (in one session) due to phantom pain and postoperative pain as a result of amputation,

\item \label{itm:fac:3} Time-varying (nonstationary) recorded biosignals, the most common causes: muscle fatigue, electrode schifts, changes of arm posture, changes of elctrode/skin impedance (sweating), doffing/donning,

\item \label{itm:fac:4} Possibility of incorrect labelling of learning objects by the amputee (reasons: misleading images of movements in control based on the paradigm of phantom motor execution with non-biomimetic (arbitrary) strategy, it is impossible to observe the movement of the amputated limb (phantom hand)),

\item \label{itm:fac:5} Poor intrasubject repeatability (changes due to human motor learning/adaptation of the user to the control of the prosthesis) – more generally denotes concept drift (different causes that could happen over the period of some time),

\item \label{itm:fac:6} Contamination of biosignals: three types of signal contaminants: (1) noise (thermal and flicker noise, amplifier saturation, analog to digital signal clipping, quantisation noise), (2) interference (power line, ECG, crosstalk), (3) artefacts (measurement artefacts, baseline wander, motion artefact, poor electrode placement),

\item \label{itm:fac:7} A large number of primary features extracted from multimodal and multichannel biosignals,

\item \label{itm:fac:8} The control system should produce a prediction of the intended movement in real time (max delay < 200 ms).
\end{enumerate}

The purpose of the paper is to develop a recognition system that will mitigate three factors (factor~\ref{itm:fac:1} limited to EMG and MMG biosignals, factor~\ref{itm:fac:6} and factor~\ref{itm:fac:7}) from those mentioned above.
The proposed method involves the use of two co-operating multiclassifier systems that have different structures and purposes. The first system is composed of one-class classifiers related to individual EMG and MMG biosignal recording channels, and its task is to recognise (detect) contaminated channels. The aim of the second multiclassifier system is to recognise the class of movement resulting from the patient's intention and represented by the recorded multichannel biosignals.
The system consists of base classifiers using the representation (extracted features) of biosignals from different channels, which allows for taking into account inter-channel dependencies between signals in the recognition process.
The system uses a negative dynamic selection mechanism, eliminating those base classifiers that are associated with biosignal channels recognised by the one-class ensemble system as being contaminated.

The main research questions that are to be answered are as follows:
\begin{itemize}
	\item \label{itm:rq:1} \textbf{RQ1} Does the proposed method perform better than the ensemble that always selects all base classifiers?

	\item \label{itm:rq:2} \textbf{RQ2} Does the proposed method perform better than a simple classifier trained on the entire feature space?

	\item \label{itm:rq:3} \textbf{RQ3} Does the proposed method perform better than the alternative approaches presented in the literature?

	\item \label{itm:rq:4} \textbf{RQ4} How do the investigated methods perform under different values of signal-to-noise ratio?

	\item \label{itm:rq:5} \textbf{RQ5} What is the impact of the number of EMG and MMG channels used to build the ensemble.
\end{itemize}

The remainder of the article is organised as follows.
In Section~\ref{sec:RelWorks} we present works related to the issue discussed in this paper.
Section~\ref{sec:ProposedMethod} contains a comprehensive description of the proposed method.
In Section~\ref{sec:ExperimentalSetup}, which concerns experimental research, we present the signalset, the acquisition method of the set, the comparative studies protocol, the extraction methods, the selection of features, and classifier models.
The results obtained are presented and discussed in Section~\ref{sec:resultsanddisc}. Finally, concluding remarks are presented in Section~\ref{sec:Conclusions}.

\section{Related Works}\label{sec:RelWorks}

The quality of the signals (MMG, EMG) is crucial in the task of controlling a bionic hand prosthetic device. When signal quality degrades, due to external (environmental) or internal factors (e.g. muscle fatigue), controlling the device becomes more difficult~\cite{Li2022,Farago2023}. Multiple solutions have been proposed in the literature to address this issue. 

The first category is to construct better hardware that is more robust to noise and artefacts. Roland et al. proposed the use of flexible insulated capacitive sensors that are less susceptible to losing contact with the skin~\cite{Roland2020}. Another approach involves utilising circular electrodes that are not affected by rotations. However, these sensors are still prone to electrode shift~\cite{Xu2020}. For MMG signals, better quality microphones or accelerometers can be used~\cite{Li2022}. For high density EMG signals, there are also methods that reconstruct signals from noise-affected signals using signals from neighbouring sensors~\cite{Farago2023}.

The next category is to filter out any artefacts that can appear at raw (timeseries) signal level. Some approaches from this group use expert knowledge to filter out predefined types of noise i.e. power line disturbances~\cite{Unanyan2019,Boyer2023}, noises comming from electrical activity of heart~\cite{Boyer2023}  or low frequency motion artefacts~\cite{Wang2019,Roland2020}. Other uses more general filtering approaches, such as multiscale PCA~\cite{Subasi2020}, or variational mode decomposition~\cite{Li2022,Ashraf2023}.

The third category consists of methods that utilise machine learning to deal with signal disturbances. The methods in this category can be subdivided into more subcategories. The first subgroup contains methods exclusively aimed at detecting disturbed signals and stopping the prosthesis from performing the movement based on faulty signals~\cite{Ding2022}. Multiclass classifiers can detect contaminated signals, but it is crucial that contaminated and labelled patterns are present in the training dataset~\cite{Irfan2023}. This kind of method can use different classifiers to detect signal contamination, i.e. SVM~\cite{Irfan2023}, or various types of neural networks~\cite{Machado2020}. However, it requires prediction of the types of noise patterns before using expert knowledge. Consequently, these types of methods can only detect certain types of contamination, while the other types remain undetected. To overcome this limitation, one-class classification can be used. This kind of machine learning is designed to detect outlier samples when the training set is composed only of uncontaminated signals. For example, the method proposed by Freaser et al. detects noisy channels using one-class SVM classifier~\cite{Fraser2014}. Contaminated signals can also be detected and used to alter the main classification system's working principle. Furukawa et al. proposed a multiclassifier system that allows one EMG channel to be contaminated~\cite{Furukawa2015}. For an eight-channel EMG signal, they trained an ensemble containing eight different classifiers. Each of them uses only seven channels. When one of the channels is contaminated, a proper seven-channel classifier is used. They also cover the case where no contamination is detected. In that case, a classifier built using all eight EMG channels is used.  Some researchers propose to retrain the entire classification system when spoilt channels are detected. However, this approach is not very practical due to the large number of retraining actions that may need to be taken~\cite{Reynolds2021}. These systems are used primarily in situations where a sensor fault is permanent. So, the retraining procedure occurs rarely. Researchers developing this kind of algorithms aim at accurate detection of faulty sensors and shorten the retraining time~\cite{Reynolds2021}. On the other hand, a classifier system that does not need explicit noisy channel detection may also be proposed. An example is to simply use a robust classification ensemble such as an error-correcting output code-based classifier~\cite{Sarabia2023}. The other is to use noise-augmented data to build a robust classification system~\cite{Lin2023}. 

In this paper, we propose a multiclassifier system with a dynamic ensemble selection strategy driven by the output of a one-class classifier-based detector of noisy signal channels. The proposed method can be seen as a generalisation of the method proposed in~\cite{Furukawa2015}.

\section{Proposed Method}\label{sec:ProposedMethod}

Let us consider a system for the control of a bionic upper limb prosthesis based on the multichannel EMG and MMG signals recognition scheme. 
Let
\begin{equation}  \label{eq_channel_set}
\mathcal{C}=\mathcal{C}_{EMG} \cup \mathcal{C}_{MMG} = \{C_1, C_2, \ldots, C_{2L}\} 
\end{equation}
denotes the set of EMG and MMG signals recorded from $2L$ sensors (channels) located on the patient's forearm stump. 
We assume the same number of channels ($L$) for both biosignals, because the EMG and MMG sensors are typically placed in pairs in a common housing evenly spaced around the forearm. 
 
The signals (\ref{eq_channel_set}) represent the patient's intention to perform a specific movement (grip or manipulation) by the prosthesis.
Control of the prosthesis in amputees is based on the phantom motor execution (PME) paradigm, i.e. the image of movement of a phantom hand ("movement without movement") that cannot be observed \cite{Garbarini2018}. Therefore, only the user knows what intention of the prosthesis movement (what class) the recorded signals represent. Moreover, the user has complete freedom in defining the meaning of classes so that they are of practical importance and will be useful in everyday life. 
For this reason, the intention of the movement (meaning of the class) is irrelevant at the classification stage. This meaning will be determined at the stage of the interpretation of the classification result, i.e. in the kinematic controller. 
Therefore, we can label classes in a general way, for example using consecutive natural numbers. 

Let then
\begin{equation}   \label{eq_classes}
\mathcal{M}=\{1,2, \ldots , M\}
\end{equation}
denotes a set of class numbers (labels). The number of classes $M$ for each user may be different, as it is related to the size of the PME repertoire and the user's ability to activate the stump muscles.   

For each channel, a feature extraction and a feature dimensionality reduction procedures are applied that collectively transform the recorded signal $C_l$ into a feature vector $x_l$ in the $d_l$ dimensional feature space $\mathcal{X}_l \subseteq \mathbb{R}^{d_l} (l=1,2,\ldots, 2L)$. Many methods and techniques have been used to extract time series features in the bioprosthesis control task with varying success \cite{MendesJunior2020}. In experimental studies (Section~\ref{sec:ExperimentalSetup}) features were extracted using the discrete wavelet transform (DWT).  

We assume that biosignals~\eqref{eq_channel_set} can be contaminated with noise during the acquisition process. Contaminations can appear randomly in different biosignal channels, so we do not know in advance how many and which channels are contaminated. Our goal is to develop a robust recognition system that will identify contaminated channels and use this information to improve the performance of classification algorithms. The proposed system consists of two co-operating ensemble classifiers, the description of which is presented below.

\subsection{Ensemble of one-class classifiers}\label{sec:ProposedMethod:oneclass}

The first multiclassifier system consists of an ensemble of base classifiers:
\begin{equation}   \label{eq_one_class_ensemble}
\Phi=\{ \phi_1, \phi_2, \ldots, \phi_{2L}\}.
\end{equation}
The base classifier $\phi_l(x_l)$ ($l=1,2, \ldots, 2L$) is one-class classifier which operates in the space $\mathcal{X}_l$ of features $x_l$ extracted from the biosignal $C_l$ registered in the $l$th channel.

The prediction of $\phi_l(x_l)$ may be zero or one. Zero indicates that $x_l$ is an outlier and the corresponding channel-specific signal $C_l$ is recognised as contaminated with some kind of noise. Consequently, the prediction equal to one indicates that $x_l$ is recognised as a target class object, i.e., the corresponding biosignal $C_l$ is considered free of contaminants. Each base classifier $\phi_l \in \Phi$ is trained using a channel-specific training set $\mathcal{T}^{\phi}_{l}$ 
that contains only objects of target class without the presence of outliers. This is the usual way to build one-class classifiers~\cite{Seliya2021}. This is because the one-class classifiers build only a model of the target class that is capable of detecting examples that do not belong to this class. In our study, target class objects are the EMG and MMG signals registered under laboratory conditions. 
Since the recording conditions are strictly controlled during the experiment, all in-laboratory registered signals are assumed to be noise/disturbance free. 
This approach is used in other work related to the detection of EMG and MMG signal contamination~\cite{Fraser2014}.

\subsection{Multiclassifier with a dynamic selection procedure}\label{sec:ProposedMethod:multiclassifier}

The second multiclassifier system:
\begin{equation}    \label{eq_ensemble}
\Psi_K = \{\psi^{(1)}_{K}, \psi^{(2)}_{K}, \ldots, \psi^{(|\Psi_K|)}_{K}\} 
\end{equation}
is designed to recognise the class of prosthesis movement from the set~\eqref{eq_classes} in accordance with the patient's intention. 
All base classifiers $\psi_K^{(k)}$ of the system (\ref{eq_ensemble}) use $K \in \{ 1,2,\ldots, 2L\} $ EMG or MMG signals from the set $\mathcal{C}$. The index $k$ is connected to the subset of $K$ signals $\mathcal{C}^{(k)}_K \subset \mathcal{C}$. We suppose that the set~\eqref{eq_ensemble} is a complete one, therefore the cardinality of the multiclassifier system $\Psi_K$ (the number of base classifiers in (\ref{eq_ensemble})) is equal to the number of $K$-combinations of the set of $2L$ channels, that is:
\begin{equation}   \label{eq_ensemble_card}
 |\Psi_K|= \binom{2L}{K} = \frac{(2L)!}{K! (2L-K)!}.
\end{equation}

The joint feature space that each base classifier $\psi^{(k)}_{K}$ operates in is defined as a Cartesian product of channel-related feature spaces:
\begin{equation}\label{eq_prod_feature_space}
\mathcal{X}^{(k)}_K  = \left(\bigtimes_{l: C_l \in \mathcal{C}^{(k)}_K} \mathcal{X}_l \right) \subseteq \mathbb{R}^{d^{(k)}_K}.
\end{equation}

In other words, vector $x^{(k)}_K$ is a concatenation of channel-specific attribute vectors:
\begin{equation}
 x^{(k)}_K = \concat_{l: C_l \in \mathcal{C}^{(k)}_K} x_l.
\end{equation}

The prediction for an object $x^{(k)}_{K} \in \mathcal{X}^{(k)}_K$ is denoted as $\psi^{(k)}_{K}(x^{(k)}_K) \in \mathcal{M}$. Each base classifier $\psi^{(k)}_{K}$ is trained using a supervised training procedure that employs a base classifier-specific training set $\mathcal{T}_{\psi^{(k)}_{K}}$~\cite{Kuncheva2014}.

Channel-specific signals that contain some amount of noise may have a negative affect on the classification quality of the classifiers. Therefore, our goal is to exclude from the ensemble base classifiers whose input space~\eqref{eq_prod_feature_space} contains attributes coming from noise-contaminated channels. The exclusion is made using responses from the one-class classifiers committee $\Phi$. The following dynamic ensemble selection rule is applied:
\begin{equation}\label{eq_selection_des}
S(\psi^{(k)}_{K}, x^{(k)}_{K}) = \ivBrack{ \prod_{l: C_l \in \mathcal{C}^{(k)}_K}  \phi_l(x_l) = 1},
\end{equation}
where $\ivBrack{\cdot}$ is the Iverson bracket operator that returns true when the logical expression inside it is true. This rule selects the ensemble members for which all channels are marked as the target class. 
In the extreme case where none of the $\psi^{(k)}_{K}$ classifiers is selected by rule $S(\psi^{(k)}_{K}, x^{(k)}_{K}) \; \; \forall \psi^{(k)}_{K} \in \Psi_K $, all the base classifiers of $\Psi_K$ are selected. The majority voting rule is used to produce the final prediction of the ensemble after selecting the base classifiers of the ensemble~\cite{Kuncheva2014}. 

\section{Experimental Setup}\label{sec:ExperimentalSetup}

The experimental study is conducted to answer the research questions posed in Section~\ref{sec:Introduction}. To do so, we compared the following methods:
\begin{itemize}
	\item \textbf{B}: A single classifier trained on the data that comes from all available EMG and MMG channels.

	\item \textbf{EC}: Error correcting output codes ensemble trained on the data that comes from all available EMG and MMG channels~\cite{Sarabia2023}. In this work, a simpler approach than that presented in~\cite{Sarabia2023} is used. The reason for this is to use a base model that is not partial least squares. We considered the following values of \textit{code\_size} parameter $\{2, 3, 4, 5, 6 \}$ The parameters are tuned with simple search approach, using three-fold cross-validation.

	\item \textbf{Or}: An oracle ensemble $\Psi_{K}$ built using given $K$-sized channel groups. This reference method tells us whether at least one of the base classifiers of the ensemble predicted the correct class for a given test sample.

	\item \textbf{Fu}: An ensemble that always selects all base classifiers from $\Psi_{K}$. This reference method tells us if the proposed dynamic committee selection procedure allows for some improvement.

	\item \textbf{DO7}: A seven channel ensemble, with a default model, which is described in~\cite{Furukawa2015}.

	\item \textbf{DO}: The proposed method.

\end{itemize}

\begin{figure}[htb]
    \centering
        \includegraphics[width=0.3\textwidth]{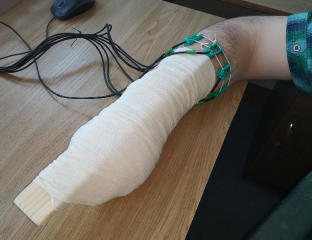}
    \caption{Illustration of  recording EMG and MMG signals. }
    \label{figs:Photo}
\end{figure}

The experiments were carried out using signals obtained from a male able-body subject, aged 73, right-handed. When creating the set of signals, the amputation was mimicked by completely immobilising the hand during the recording of the signals, which obliges the subject to generate EMG/EMG signals based on his own image of the movement performed, not on the basis of actual movements performed by the hand~\cite{Schone2023,Kristoffersen2020}.
The biosignals were registered using a special designed measuring circuit with 4 EMG sensors and 4 MMG sensors (L=4) evenly spaced around the forearm and a sampling frequency of 1000 Hz. The detailed description of the measuring circuit is described in~\cite{Wolczowski2017}.  The signalset used in the experiments consisted of eight classes (according to the subject's imagination): (1) wrist flexion, (2) wrist extension, (3) ulnar deviation, (4) radial deviation, (5) index and middle fingers flexion, (6) index and middle fingers extension, (7) ring and little fingers flexion, (8) ring and little fingers extension. One hundred four-channel EMG and four-channel MMG signals per class were recorded. Each measurement lasted 1000 ms and was preceded with a 3 s break. Figure~\ref{figs:Photo} shows the placements of the sensor and the hand immobilisation technique. The subject gave his informed consent prior to participation.

For the proposed methods, the Random Forest classifier is used as the base classifier. The size of the Random Forest Ensemble is set to 30. The ensembles are homogeneous. It means that for the ensemble, only one classification algorithm (Random Forest) is used as the base classifier. As an outlier detector, we use a one-class SVM classifier. This classifier proved to be a useful tool in the task of detecting contaminated EMG signals~\cite{Fraser2014}. The $\nu$ parameter of this classifier is tuned using the following procedure. Using a threefold cross-validation procedure, the training and validation sets are extracted. The validation (testing) part is then augmented with artificial examples marked as noisy ones. Artificial examples are generated using a uniform distribution in the classifier-specific input space $\mathcal{X}_l$. The following values of $\nu$ are considered $\{ 0.1, 0.2, \ldots, 1.0\}$. The classification quality is assessed using the balanced accuracy criterion (binary classification task). When the best value of $\nu$ is found, the final one-class classifier is trained using the entire training set $\mathcal{T}^{\phi}_{l}$.  We use the base classifiers implemented in the scikit-learn package. Unless otherwise specified, the classifier parameters are set to their default values.

To simulate real-world EMG and MMG signal contaminations, the following noise generation techniques are used~\cite{Moura2018, Wang2019,Farago2023,Boyer2023}:
\begin{itemize}

	\item Simulation of power grid noise by injection of a sinusoidal signal whose frequency varies from 48 to 52 Hz. The amplitude of the inserted noise depends on the SNR rate considered in the experimental scenario. This kind of noise may affect both EMG and MMG signals.

	\item Signal attenuation that simulates the sensor losing contact with the skin. For MMG signals, it simulates the situation when the distance between the microphone and the skin increases. The attenuation level depends on the selected SNR.

	\item Gaussian noise. It simulates the general noises that may appear in the signal acquisition circuit. This kind of noise may affect both EMG and MMG signals.

	\item Simulation of non-linear amplifier characteristics for signals of high amplitude. This is done by non-linear clipping the peaks of the signal. This kind of noise may affect both EMG and MMG signals. The clipping level depends on the selected SNR. 

	\item Baseline wandering. This is simulated by injecting low-frequency (0.5 to 1.5 Hz) sinusoidal noise. The amplitude of the signal depends on the selected SNR. For MMG signals, frequencies coming from this interval simulate motion artefacts. 

\end{itemize}
In experimental studies, we consider the following SNR levels $\{ 0,1,2, \ldots, 6, 10 \}$.

The training and testing sets are obtained by a ten-fold cross-validation repeated 4 times. The testing dataset is then randomly contaminated with one of the above-mentioned noise types with a selected SNR level. The number of contaminated channels is also randomly selected from the following set of values $\{1, 2, \ldots, 7 \}$. The number of channels used to create base classifiers of the $\Psi_{K}$ ensemble is $K \in \{ 1, 2, 3, \ldots, 7, \{2, 3, 4\}, \{3,4, 5\}, \{2, 3, 5\}, \{2, 4, 5\} \}$. In the above-mentioned set, subsets like $\{2,3,4\}$ mean that the ensemble contains all base classifiers for the mentioned numbers of channels. In other words, a joint ensemble is built that aggregates the base classifiers coming from the ensembles built for the considered values of $K$.

Feature vectors were created from raw EMG and MMG signals using the discrete wavelet transform technique. The \textit{db6} wavelet and three levels of decomposition were used. The following functions were calculated for the transformation coefficients \cite{MendesJunior2020}: MAV (mean absolute value), SSC (slope sign change). Due to the relatively small sample bot experiments were performed on the same dataset.

Following the recommendations of~\cite{garcia2008extension}, the statistical significance of the results obtained was evaluated using the pairwise Wilcoxon signed rank test. Since multiple comparisons were made, family-wise errors (FWER) should be controlled. To do so, Holm's procedure of controlling the FWER was employed. For all tests, the significance level was set to $\alpha=0.05$. We used statistical tests and correction procedures implemented in SciPy. For some analysis, the average rank approach is also used. To assess the classification quality, the balanced accuracy criterion is employed.
The experimental code is provided in~\footnote{\url{https://github.com/ptrajdos/CLDD_2024.git}}

\section{Results and discussion}\label{sec:resultsanddisc}

\begin{figure}[htb]
    \centering
        \includegraphics[width=0.4\textwidth]{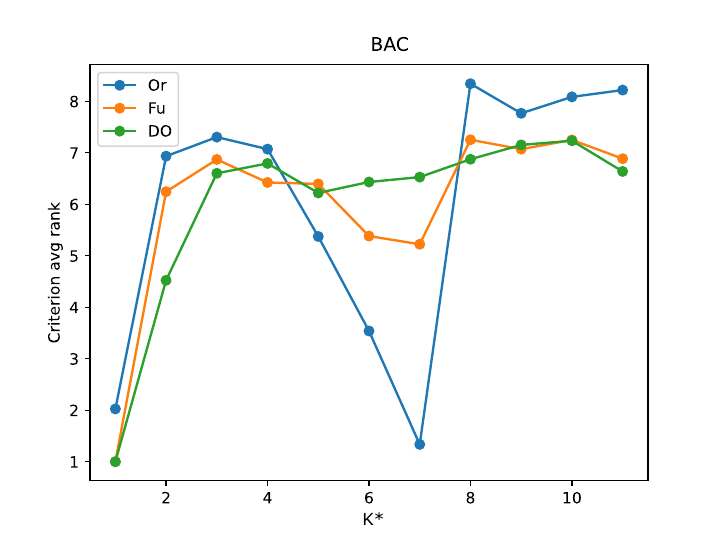}
    \caption{The impact of changing $K$, signal with noise -- ranks.}
    \label{figs:gs_ranks}
\end{figure}

\begin{figure}[htb]
    \centering
        \includegraphics[width=0.4\textwidth]{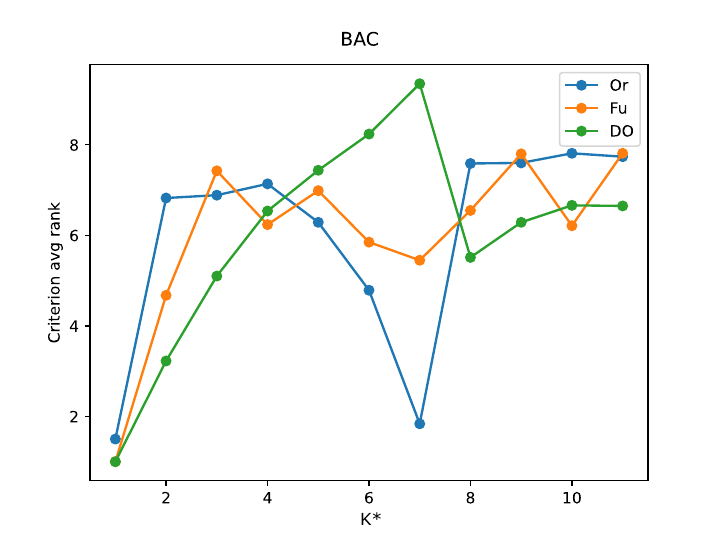}
    \caption{The impact of changing $K$, signal without noise -- ranks.}
    \label{figs:gs_ranks_no_noise}
\end{figure}

Let us begin with an experiment that examines how changing $K$ affects the classification quality of the ensemble methods. During the experiment, the following values are investigated: $K \in \{ 1, 2, 3, \ldots, 7, \{2, 3, 4\}, \{3,4, 5\}, \{2, 3, 5\}, \{2, 4, 5\} \}$. In Figures~\ref{figs:gs_ranks},~\ref{figs:gs_ranks_no_noise}  and Tables~\ref{table:gs_ranks},~\ref{table:gs_ranks_no_noise}  the above-mentioned values of $K$ are denoted using natural numbers $\{1,2,\ldots, 11\}$. If $K$ is represented as a set, it means that we investigate the joint committee created using the mentioned values of $K$. Figures~\ref{figs:gs_ranks} and \ref{figs:gs_ranks_no_noise} present the averaged ranks (averaged over different SNR values and different folds) related to different $K$ values for each investigated ensemble method. In other words, for each method the values of $K$ are ranked. Tables~\ref{table:gs_ranks} and \ref{table:gs_ranks_no_noise} present the average ranks and the outcome of the statistical test performed. Each row of the table presents the average ranks related to an ensemble method, the name of which is presented in the first column. Each of the remaining columns is related to different values of $K$ used to create the ensemble. The subscript numbers below the average rank value present the column numbers for the methods to which the method with the selected value of $K$ is significantly better. When there are no significant differences between the column of interest and the other columns, a pause sign `--' is placed in the subscript.

 We start with experiments conducted on mixture of clean and noisy signals. It is evident from the Figure~\ref{figs:gs_ranks} and the Table~\ref{table:gs_ranks} that the classification quality obtained by each of the ensemble methods is strongly influenced by $K$. Compared to other values of $K$, all ensemble methods perform poorly for $K=1$ (one classifier for one EMG or MMG channel). This is because different muscles are activated in a different way when the subject generates different contraction patterns. Consequently, a lot of information about the muscle activation performed is encoded in differences between channels~\cite{Trajdos2023}. Even for $K=2$, the increase in classification quality is significant compared to $K=1$. An interesting result may be observed for the oracle classifier (\textbf{Or}) when ensembles are built only for a single value of $K$. The best performance (in terms of average ranks) is observed when $K=3$. However, there are no significant differences between $K\in \{ 2,3,4\}$. When the value of $K$ increases over 4, the average rank decreases significantly. Since the oracle algorithm selects the correct class if one is presented in the ensemble, the drop in classification quality means that the diversity of the ensemble is the highest for $K \in{2,3,4}$ and then drops. The diversity is then clearly related to the size of the ensemble $|\Psi_{K}|$. However, the size of the ensemble is not the only factor that affects the performance of the classifiers. If the ensemble size was only factor, then, for a single value $K$, the best performance should be observed for $K=4$ since the ensemble size is the highest for this value. When the ensemble is built using more signal channels, then the base classifiers become less diverse since they share more channel-specific information.  When multiple values of $K$ are used to build a joint ensemble, the performance of the oracle classifier is also high due to the high diversity. A similar pattern of average ranks can be observed for the ensemble that always selects all base classifier \textbf{Fu}. For this type of ensemble, diversity also plays an important role, as the best results are obtained for committees containing a relatively large number of classifiers $ K \in \{ 2,3,4,5\}$. For the proposed method, on the other hand, no significant differences can be observed for $K \in \{ 3,4,5,6,7\}$. For those values of $K$, the differences between the average ranks are also smaller and do not decrease if $K$ increases. This means that even if the committee is smaller and less diverse, the outlier selection procedure successfully eliminates the base classifiers related to noisy channels. Consequently, the classification quality does not drop.

 Now, we analyse the situation where there is almost no artificial noise (contamination) in the test signals (SNR=10). The results are presented in Figure~\ref{figs:gs_ranks_no_noise} and Table~\ref{table:gs_ranks_no_noise}. The behaviour of the \textbf{Or} and \textbf{Fu} methods is quite similar when there is no noise present. That is, the methods perform better when the size and diversity of the ensemble is high. Joining the committees built for different values of $K$ has a positive impact on the quality of classification. The behaviour of the proposed method \textbf{DO}, on the other hand, changes significantly. That is, the classification quality continues to increase when higher values of $K$ are considered. Even if the signals are not contaminated with artificial noise, the removal of some base classifiers from the committee allows the classification quality to be increased. This may mean that even signals registered under strictly controlled laboratory conditions may contain some degree of noise that is successfully detected by the outlier detectors. The quality of the classification can be improved by removing the slightly contaminated channels. The increase in classification quality is also related to the number of channels used to build each of the base classifiers. The more channels we include, the better the classification quality. This is due to the fact that important information about muscle activation pattern is encoded in differences between signals registered for different muscles~\cite{Trajdos2023}. Even combining a larger number of simpler base classifiers cannot outperform the results obtained by base classifiers built for a larger number of channels. However, as the above-described results show, this effect holds only for signals whose contamination level is relatively low. For signals with injected artificial noise, more diverse ensembles are preferable. Taking this into account, the joint ensembles that combine different values of $K$ should combine smaller and larger values of $K$ to improve the classification quality of clean signals.

 Based on the results, it is recommended to establish joint committees for various values of $K$. It is clear that for all the multiclassifier systems considered, the joint committees offer the best classification quality. For the second analysis performed, we use the build committees for $K=7$ (to provide a better comparison with the reference method \textbf{DO7}) and $K=\{ 2,3,5\}$ to show the abilities of the joint committees built for different values of $K$.

{
\setlength{\tabcolsep}{3pt}
\begin{table*}[htb]
\renewcommand{\arraystretch}{0.6}
\centering
\footnotesize
\caption{The impact of changing $K$, signal with noise -- statistical tests.\label{table:gs_ranks}}
\begin{tabular}{llllllllllll}
 & \multicolumn{11}{c}{BAC} \\
 & 1 & 2 & 3 & 4 & 5 & 6 & 7 & 8 & 9 & 10 & 11 \\
 \cmidrule(lr){2-12}
Or & 2.025 & 6.934 & 7.305 & 7.072 & 5.375 & 3.539 & 1.336 & 8.342 & 7.767 & 8.086 & 8.219 \\
 & {\tiny 7} & {\tiny 1,5,6,7} & {\tiny 1,5,6,7} & {\tiny 1,5,6,7} & {\tiny 1,6,7} & {\tiny 1,7} & -- & {\tiny 1,2,3,4,5,6,7,9} & {\tiny 1,2,3,4,5,6,7} & {\tiny 1,2,3,4,5,6,7,9} & {\tiny 1,2,3,4,5,6,7,9} \\
Fu & 1.000 & 6.245 & 6.870 & 6.423 & 6.395 & 5.383 & 5.223 & 7.253 & 7.070 & 7.250 & 6.886 \\
 & -- & {\tiny 1,6,7} & {\tiny 1,6,7} & {\tiny 1,6,7} & {\tiny 1,6,7} & {\tiny 1} & {\tiny 1} & {\tiny 1,2,4,5,6,7} & {\tiny 1,2,6,7} & {\tiny 1,2,4,5,6,7} & {\tiny 1,6,7} \\
DO & 1.000 & 4.523 & 6.602 & 6.794 & 6.220 & 6.433 & 6.528 & 6.875 & 7.153 & 7.234 & 6.638 \\
 & -- & {\tiny 1} & {\tiny 1,2} & {\tiny 1,2} & {\tiny 1,2} & {\tiny 1,2} & {\tiny 1,2} & {\tiny 1,2} & {\tiny 1,2,3,5,6,7} & {\tiny 1,2,3,5,6,7,11} & {\tiny 1,2} \\
\end{tabular}
\end{table*}
}


{
\setlength{\tabcolsep}{3pt}
\begin{table*}[htb]
\renewcommand{\arraystretch}{0.6}
\centering
\footnotesize
\caption{The impact of changing $K$, signal without noise -- statistical tests.\label{table:gs_ranks_no_noise}}
\begin{tabular}{llllllllllll}
 & \multicolumn{11}{c}{BAC} \\
 & 1 & 2 & 3 & 4 & 5 & 6 & 7 & 8 & 9 & 10 & 11 \\
 \cmidrule(lr){2-12}
Or & 1.500 & 6.825 & 6.888 & 7.138 & 6.287 & 4.787 & 1.837 & 7.588 & 7.600 & 7.812 & 7.737 \\
& -- & {\tiny 1,6,7} & {\tiny 1,6,7} & {\tiny 1,6,7} & {\tiny 1,7} & {\tiny 1,7} & -- & {\tiny 1,6,7} & {\tiny 1,6,7} & {\tiny 1,5,6,7} & {\tiny 1,6,7} \\
Fu & 1.000 & 4.675 & 7.425 & 6.237 & 6.987 & 5.850 & 5.450 & 6.550 & 7.800 & 6.213 & 7.812 \\
 & -- & {\tiny 1} & {\tiny 1,2} & {\tiny 1} & {\tiny 1,2} & {\tiny 1} & {\tiny 1} & {\tiny 1,2} & {\tiny 1,2} & {\tiny 1} & {\tiny 1,2,7} \\
DO & 1.000 & 3.225 & 5.100 & 6.537 & 7.438 & 8.238 & 9.350 & 5.513 & 6.287 & 6.662 & 6.650 \\
 & -- & {\tiny 1} & {\tiny 1,2} & {\tiny 1,2} & {\tiny 1,2,3} & {\tiny 1,2,3,8,9} & {\tiny 1,2,3,4,5,8,9,10,11} & {\tiny 1,2} & {\tiny 1,2} & {\tiny 1,2} & {\tiny 1,2} \\
\end{tabular}
\end{table*}
}

\begin{figure}[htb]
    \centering
        \includegraphics[width=0.4\textwidth]{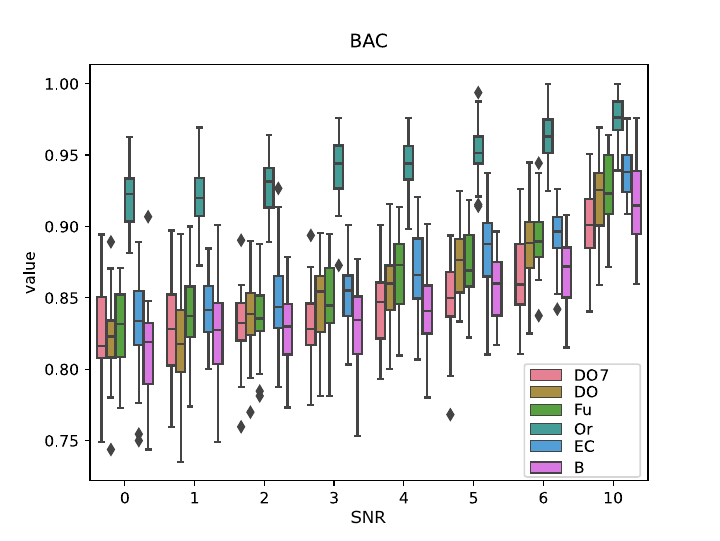}
    \caption{The impact of changing SNR, $K=7$ -- boxplots.\label{fig:snr_gs_7}}
\end{figure}

\begin{figure}[htb]
    \centering
        \includegraphics[width=0.4\textwidth]{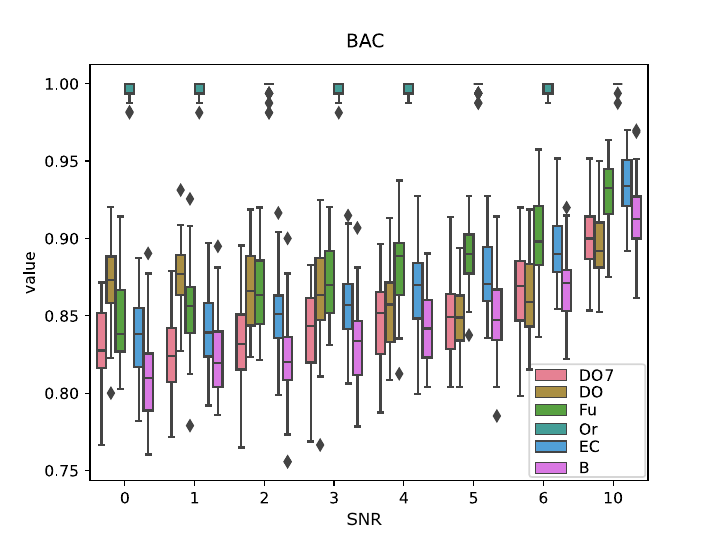}
    \caption{The impact of changing SNR, $K \in \{2, 3, 5\}$ -- boxplots.\label{fig:snr_gs_10}}
\end{figure}

The results related to different SNR values are presented in Figures~\ref{fig:snr_gs_7} -- \ref{fig:snr_gs_10} and Tables~\ref{table:snr_gs_7} -- \ref{table:snr_gs_235}. Figures present method-related boxplots for different values of SNR. Box plots are colour encoded. The tables present the method-related average (over different folds) ranks for different values of SNR. Each row of a table is related to different SNR values. Each column is related to a different method. Subscripts related below each average ranks contain indices of methods (number of columns) that given method is significantly better.

Let us start with the results related to $K=7$. As shown in Figure~\ref{fig:snr_gs_7} and Table~\ref{table:snr_gs_235}, it is evident that the classification quality of all investigated methods increases with increasing SNR. This is clear because higher values of SNR mean that less noise is present in the signals. The oracle classifier is undoubtedly the classifier with the highest-quality classification. This classifier is a perfect committee selection procedure that possesses the information about the true class of the object. Its superior performance over \textbf{B} and \textbf{EC} shows that the proposed method of ensemble creation has a great potential that can be revealed by a proper committee selection technique. On the other hand,~\textbf{Fu} represents the most naive committee selection technique that always uses the full ensemble to make a prediction. Despite of its simplicity~\textbf{Fu} method in some cases outperforms~\textbf{B} method. This is further confirmation that the proposed method of committee creation has a great potential, since it is outperformed only by~\textbf{EC} for $\mathrm{SNR}=10$. However, a committee selection procedure plays the most important role since an improper selection procedure can worsen the classification quality. For our experiment, the selection rule used by~\textbf{DO7} does not perform very well. It is unable to outperform any other of the investigated methods and, more importantly, it is often outperformed by~\textbf{Fu}, \textbf{DO} and \textbf{EC}. This is due to the \textbf{DO7} method being designed in such a way that it always selects only one classifier from the ensemble. The selected classifier may be a classifier trained on the entire dataset (no-noise detected by one-class classifiers) or a dateset with one channel removed. The assumption that there can be only one noisy channel is overoptimistic, since in real-world data, more channels may be noisy. This is why our more general method seems to perform better than~\textbf{DO7}. The committee selection method proposed in this paper (\textbf{DO}) is often better than \textbf{B} and, in general, is comparable to \textbf{Fu} and \textbf{EC}. Being comparable to \textbf{Fu} is not a great advantage, since the proposed method uses an additional one-class committee that results in a higher computational burden.

{
\setlength{\tabcolsep}{3pt}
\begin{table}[htb]
\renewcommand{\arraystretch}{0.6}
\centering
\footnotesize
\caption{The impact of changing SNR, $K=7$ -- statistical tests.\label{table:snr_gs_7}}

\begin{tabular}{lllllll}
 & \multicolumn{6}{c}{BAC} \\
 & B & EC & Or & Fu & DO & DO7 \\
\cmidrule(lr){2-7}

SNR:10 & 2.788 & 4.188 & 5.875 & 3.375 & 3.038 & 1.738 \\
 & {\tiny 6} & {\tiny 1,4,5,6} & {\tiny 1,2,4,5,6} & {\tiny 6} & {\tiny 6} & -- \\
SNR:6 & 1.988 & 3.862 & 6.000 & 3.550 & 3.300 & 2.300 \\
 & -- & {\tiny 1,6} & {\tiny 1,2,4,5,6} & {\tiny 1,6} & {\tiny 1,6} & -- \\
SNR:5 & 2.350 & 3.962 & 6.000 & 3.475 & 3.288 & 1.925 \\
 & -- & {\tiny 1,6} & {\tiny 1,2,4,5,6} & {\tiny 1,6} & {\tiny 1,6} & -- \\
SNR:4 & 2.263 & 3.712 & 6.000 & 3.638 & 3.163 & 2.225 \\
 & -- & {\tiny 1,6} & {\tiny 1,2,4,5,6} & {\tiny 1,6} & {\tiny 1,6} & -- \\
SNR:3 & 2.325 & 3.750 & 6.000 & 3.237 & 3.288 & 2.400 \\
 & -- & {\tiny 1,6} & {\tiny 1,2,4,5,6} & {\tiny 1,6} & {\tiny 1,6} & -- \\
SNR:2 & 2.562 & 3.725 & 6.000 & 2.888 & 3.188 & 2.638 \\
 & -- & {\tiny 1,6} & {\tiny 1,2,4,5,6} & -- & -- & -- \\
SNR:1 & 2.663 & 3.587 & 5.987 & 3.450 & 2.513 & 2.800 \\
 & -- & {\tiny 1,5} & {\tiny 1,2,4,5,6} & {\tiny 5} & -- & -- \\
SNR:0 & 2.612 & 3.425 & 6.000 & 3.175 & 2.712 & 3.075 \\
 & -- & {\tiny 1} & {\tiny 1,2,4,5,6} & -- & -- & -- \\
\end{tabular}

\end{table}
}

Fortunately, the situation changes when the size of the initial committee and its diversification increase. This case occurs for $K \in \{2,3,5\}$. For this experiment, the~\textbf{Or} classifier achieved almost constant classification quality, approximately equal to $1.0$. This means that with proper committee selection procedures, the potential of the committee is great. The naive selection procedure used by~\textbf{Fu} also allows it to be one of the best performing methods in this comparison, as it outperforms~\textbf{B} and~\textbf{OD7}. It is also comparable to~\textbf{EC}. In this experimental scenario, the proposed method exhibits a different behaviour. For low SNR values (0, 1), it outperforms \textbf{B}, \textbf{EC}, \textbf{Fu} and \textbf{DO7}. This means that for low SNR values, the removal of noisy channels offers a great improvement. However, the positive differences between \textbf{DO} and remaining methods disappear when the SNR increases. For high SNR values, the proposed method is outperformed by \textbf{EC} and \textbf{Fu} (for $\mathrm{SNR}=10$ also by \textbf{B}). This leads to the conclusion that the complete removal of noisy channels is beneficial for a large amount of noise in these channels. However, when the amount of noise is low, the base classifiers of the ensemble are robust enough to deal with this noise. Consequently, the selection of all classifiers within the ensemble offers better classification quality. Therefore, in future work, a new method of soft committee selection should be developed using the above-mentioned observation in practice.

{
\setlength{\tabcolsep}{3pt}
\begin{table}[htb]
\renewcommand{\arraystretch}{0.6}
\centering
\footnotesize
\caption{The impact of changing SNR, $K \in \{2, 3, 5\}$ -- statistical tests.\label{table:snr_gs_235}}

\begin{tabular}{lllllll}
 & \multicolumn{6}{c}{BAC} \\
  & B & EC & Or & Fu & DO & DO7 \\
\cmidrule(lr){2-7}

SNR:10 & 2.788 & 4.287 & 6.000 & 3.825 & 1.875 & 2.225 \\
 & {\tiny 5} & {\tiny 1,5,6} & {\tiny 1,2,4,5,6} & {\tiny 1,5,6} & -- & -- \\
SNR:6 & 2.237 & 3.875 & 6.000 & 4.300 & 2.112 & 2.475 \\
 & -- & {\tiny 1,5,6} & {\tiny 1,2,4,5,6} & {\tiny 1,5,6} & -- & -- \\
SNR:5 & 2.225 & 3.600 & 6.000 & 4.400 & 2.350 & 2.425 \\
 & -- & {\tiny 1,5,6} & {\tiny 1,2,4,5,6} & {\tiny 1,2,5,6} & -- & -- \\
SNR:4 & 2.062 & 3.425 & 6.000 & 4.225 & 2.862 & 2.425 \\
 & -- & {\tiny 1,5,6} & {\tiny 1,2,4,5,6} & {\tiny 1,2,5,6} & -- & -- \\
SNR:3 & 1.750 & 3.087 & 6.000 & 4.025 & 3.737 & 2.400 \\
 & -- & {\tiny 1,6} & {\tiny 1,2,4,5,6} & {\tiny 1,2,6} & {\tiny 1,6} & -- \\
SNR:2 & 1.762 & 3.013 & 6.000 & 3.800 & 4.175 & 2.250 \\
 & -- & {\tiny 1,6} & {\tiny 1,2,4,5,6} & {\tiny 1,2,6} & {\tiny 1,2,6} & -- \\
SNR:1 & 1.950 & 2.888 & 6.000 & 3.538 & 4.588 & 2.038 \\
 & -- & {\tiny 1,6} & {\tiny 1,2,4,5,6} & {\tiny 1,6} & {\tiny 1,2,4,6} & -- \\
SNR:0 & 1.637 & 2.875 & 6.000 & 3.475 & 4.475 & 2.538 \\
 & -- & {\tiny 1} & {\tiny 1,2,4,5,6} & {\tiny 1,6} & {\tiny 1,2,4,6} & {\tiny 1} \\

\end{tabular}

\end{table}
}

To summarise, the answers to the research questions are as follows:
\begin{itemize}
	\item \textbf{RQ1} In general, the proposed method \textbf{DO} is comparable to \textbf{Fu}. In case of low SNR the proposed method is better.

	\item \textbf{RQ2} The proposed \textbf{DO} method performs better than \textbf{B} classifier when the amount of noise in signals is relatively high. When the amount of noise is low, then the methods are comparable.

	\item \textbf{RQ3} The proposed \textbf{DO} method performs better than \textbf{DO7} classifier when the amount of noise in signals is relatively high. When the amount of noise is low, then the methods are comparable.

	\item \textbf{RQ4} In general, the higher value of SNR is, the better the investigated methods perform. However, the proposed method shows quite stable classification quality results for low SNR values.

	\item \textbf{RQ5} In general, classification quality is more related to the ensemble size and diversity rather than the number of the included channels. However, when no noise is present, the greater number of channels (parameter $K$) is, the classification quality is better.
\end{itemize}

\section{Conclusions}\label{sec:Conclusions}

In this paper, we propose a dual ensemble multiclassifier system that is designed to deal with noise/artefacts in multimodal data consisting of EMG and MMG signals. The first ensemble system is a one-class classifiers ensemble responsible for detecting noisy/contaminated channels. A one-class classifier ensemble is used to allow the ensemble to recognise any kind of outlying signal. The output of a one-class ensemble is then used to perform dynamic ensemble selection in the second multiclassifier system. The second multiclassifier system contains base classifiers built using signals coming from multiple channels. The number of channels used to create the base classifier for this ensemble is a parameter of the system.

We performed two experiments. The first aims at determining the best value of the above-mentioned parameter $K$. The goal is also to analyse the impact of the ensemble size and diversity on the classification quality of the ensembles. The first experiment shows that, in the presence of noise, ensemble classifiers tend to provide higher quality predictions when the diversity of the ensemble is higher. It also shows that it is desirable to join multiple committees built for different values of the parameter $K$ since the diversity of such committees is greater than an ensemble built for a single value of $K$.

During the second experiment, we analyse how the proposed method performs when the signal-to-noise ratio (SNR) changes. In this experiment, we also compare the proposed method with selected reference methods. The results show that with proper settings of the $K$ parameter, the proposed method can outperform the reference methods for low SNR values (when the signals are noisy). However, the superiority of the proposed method no longer holds for higher SNR values. This is due to the partial robustness of the base classifiers (of $\Psi_{K}$) to noise. For low-noise scenarios, it is better to select all base classifiers from the ensemble. 

These observations lead us to the conclusion that a new, better dynamic ensemble selection procedure may be constructed. This new strategy should completely remove ensemble classifiers that use channels with a high level of contamination, but when the contamination degree is low, it should allow the outcome of such classifiers to be used. Because of this, our future work is aimed at developing such an ensemble selection procedure. We think that a soft ensemble selection strategy should be suitable and meets the properties mentioned above.

\begin{ack}
This work is supported by the National Center for Research and Development ({www.ncbr.gov.pl}) through project no. \textbf{ /0018/2020-00} within\textit{''Things are for people''} programme. The authors have no conflict of interest to declare. The authors thank Dr. Andrzej Wolczowski for valuable discussions, Dr. Jerzy Witkowski for the design and manufacture of EMG and MMG sensors and Dr. Michal Bledowski for the application for recording EMG and MMG signals \cite{Wolczowski2017}.
\end{ack}



\FloatBarrier
\clearpage
\bibliography{bibliography}

\end{document}